\pdfoutput=1
\documentclass[10pt,twocolumn,letterpaper]{article}

\usepackage[pagenumbers]{cvpr} 

\definecolor{cvprblue}{rgb}{0.21,0.49,0.74}
\usepackage[pagebackref,breaklinks,colorlinks,allcolors=cvprblue]{hyperref}

\def\confName{CVPR}
\def\confYear{2027}

\usepackage{hyperref}
\usepackage{url}
\usepackage[nolist]{acronym}
\usepackage{xspace}

\usepackage{makecell}
\usepackage{pdflscape}
\usepackage{adjustbox}
\usepackage{enumitem}
\usepackage{multirow}
\usepackage{pifont}
\usepackage{tabularx}
\usepackage{tablefootnote}
\usepackage{booktabs}
\usepackage{makecell}
\usepackage{wrapfig}
\usepackage{xspace}

\usepackage{amsmath,amssymb,amsthm}

\usepackage{todonotes}
 
\newcommand{\cmark}{\ding{51}}
\newcommand{\xmark}{\ding{55}}

\title{Forensic Twins: Self-Supervised Residual Learning for \\ AI-Generated Image Forensics}

\author{%
  Javier Muñoz-Haro \quad Ruben Tolosana \quad Ruben Vera-Rodriguez\\
  Aythami Morales \quad Julian Fierrez\\[4pt]
  BiometricsAI Lab, Universidad Autónoma de Madrid, Spain\\
  {\tt\small \{javier.munoz, ruben.tolosana, ruben.vera, aythami.morales, julian.fierrez\}@uam.es}
}

\newcommand{\method}{\textit{Forensic Twins}\xspace}

\begin{acronym}
    \acro{ood}[OOD]{Out-of-Distribution Detection}
    \acro{ssl}[SSL]{Self-Supervised Learning}
    \acro{ssrl}[SSRL]{Self-Supervised Residual Learning}
    \acro{tto}[TTO]{Test-Time Optimization}
    
    \acro{cnn}[CNN]{Convolutional Neural Network}
    \acro{vit}[ViT]{Vision Transformer}
    \acro{gan}[GAN]{Generative Adversarial Network}
    \acro{lora}[LoRA]{Low-Rank Adaptation}
    \acro{dm}[DM]{Diffusion Model}
    \acro{uld}[ULD]{Unconditional Latent Diffusion}
    \acro{ldm}[LDM]{Latent Diffusion Model}
    \acro{sam2}[SAM-2]{Segment Anything Model 2}
    \acro{pca}[PCA]{Principal Component Analysis}
    \acro{gmm}[GMM]{Gaussian Mixture Model}
    \acro{std}[STD]{Global Standard Deviation Pooling}
    \acro{gap}[GAP]{Global Average Pooling}
    \acro{moe}[MoE]{Mixture of Experts}
    \acro{mhsa}[MHSA]{Multi-Head Self Attention}
    \acro{fre}[FRE]{Forensic Residual Extractor}

    \acro{bt}[BT]{Barlow Twins}
    \acro{moco}[MoCo]{Momentum Contrast}
    \acro{dino}[DINO]{Self DIstillation with NO Labels}
    
    \acro{eer}[EER]{Equal Error Rate}

    \acro{gimp}[GIMP]{GNU Image Manipulation Program}
    \acro{roi}[ROI]{Region Of Interest}
    
\end{acronym}

\begin{document}
\maketitle
\begin{abstract}
Detectors of AI-generated images are typically trained using samples from specific Generative AI architectures and degrade as soon as a new architecture emerges. Recent approaches have explored self-supervised pre-training as an alternative solution, yet standard frameworks work against the forensic task, since their augmentations overwrite the micro-statistics of image formation. This paper introduces \method, a Self-Supervised Residual Learning (SSRL) framework whose pretext task removes shared image content between views. Each image is mapped through a forensic residual extractor, from which two spatially disjoint crops are drawn. Sharing no pixel, the two views leave the redundancy-reduction objective with a predominantly content-independent common signal, the stationary fingerprint of the image acquisition pipeline. The encoder is exclusively trained on real images; no AI-generated image is used to learn the representation. Experiments show competitive results in zero-shot, few-shot, and linear probing scenarios, reaching 97.21\% AUC across 26 unseen AI generators, including GANs, diffusion models, and commercial systems. We also provide insightful analysis on the learned representation regarding unsupervised clustering on different AI generators and its forensic properties. Code, weights and exact splits are publicly available\footnote{https://github.com/BiometricsAI/ForensicTwins}. 
\end{abstract}    
\section{Introduction}
\label{sec:intro}

Generative models now synthesize images with unprecedented detail \cite{diffmodels_surv, sauer2022stylegan, alaluf2022third, tolosana2020deepfakes} and increasing physical and semantic coherence \cite{Avrahami_2022_CVPR}, so that realistic synthetic media can be deployed at scale. Condition-guided generation, from class labels \cite{dhariwal2021diffusion}, text prompts \cite{Rombach_2022_CVPR}, or reference images \cite{reface}, further allows them to replicate structural layouts \cite{freemorph, diffmorph, chen2023textdiffuser} or emulate artistic styles \cite{snapshot}.

Supervised detectors of AI-generated images \cite{npr2024, wang2020cnn, ojha2023towards} degrade in \ac{ood} scenarios \cite{ai_gen_det_perf_comp} and fail to generalize to unseen generators: handcrafted forensic priors and binary objectives push them to overfit to source-specific artifacts. Instead, recent AI-generation detection methods \textit{i)} treat semantics and forensic traces as decorrelated spaces \cite{yan2024effort, guo2026omniaid}, or \textit{ii)} model the manifold \cite{zhang2025detecting} or descriptors \cite{self-descriptors-cvpr} of natural images, relying on massively parameterized backbones such as DINOv2 \cite{dinov2} and CLIP \cite{radford2021learning}, which leverage AI-generated images during training \cite{yan2024effort, guo2026omniaid} or on costly per-image \ac{tto} \cite{self-descriptors-cvpr}.

\ac{ssl} \cite{ssl_survey} is a compelling alternative, learning generalizable representations from unannotated data without costly and biased labeling. Yet standard frameworks such as \ac{bt} \cite{bt}, \ac{moco} \cite{moco, mocov2} and \ac{dino} \cite{dinov1, dinov2, dinov3} are not suited to forensics: aggressive augmentations and overlapping views let the learning process discard the low-amplitude micro-statistical noise intrinsic to natural image formation \cite{wang2020cnn, marra2019gans, sinitsa2024deep}.

\begin{figure*}[t]
    \centering
    \includegraphics[width=\textwidth]{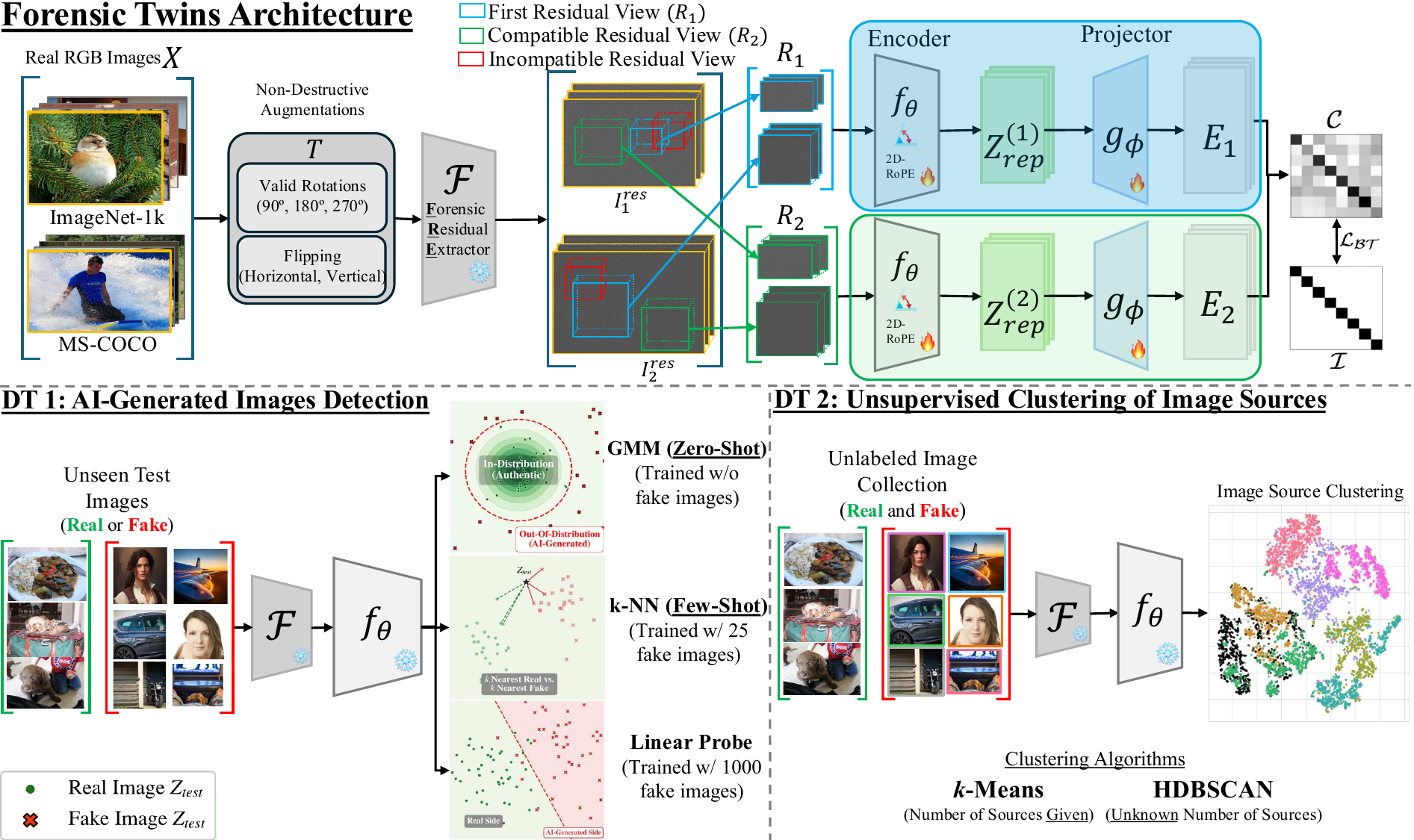}
    \caption{\textbf{Overview of \method and applications.} \textbf{Top:} Real RGB images ($X$) undergo non-destructive augmentations ($\mathcal{T}$) and residual filtering ($\mathcal{F}$). Then, two strictly disjoint crops (compatible views $R_1, R_2$) pass through a ViT ($f_\theta$) and a projector ($g_\phi$), trained with a redundancy-reduction objective on their cross-correlation matrix $\mathcal{C}$. \textbf{Bottom-left:} Different variants (zero-, few-shot, and linear probing) for AI-Generated image detection. \textbf{Bottom-right:} \textit{k}-Means and HDBSCAN groups the frozen embeddings of an unlabeled real/AI-generated collection by image source.}
    \label{fig:ft_overview}
    \vspace{-7pt}
\end{figure*}

Here we propose \method, a \ac{ssrl} framework that suppresses semantic content in favor of forensic traces. Fig.~\ref{fig:ft_overview} provides an overview of the proposed method and its applications. Following the redundancy-reduction principle of \ac{bt} \cite{bt}, \method is designed so that the low-level fingerprint of image formation is the dominant factor the objective can exploit. A frozen pre-trained forensic residual extractor \cite{self-descriptors-cvpr} suppresses macroscopic content while preserving high-frequency residuals, and two \textit{disjoint} crops of these residuals, which share no pixel and therefore cannot be aligned by matching local patterns, serve as the two views. A \ac{vit} encodes both, and driving the cross-correlation matrix of their embeddings towards the identity enforces cross-view invariance while decorrelating feature dimensions. We argue that the dominant signal both crops still share is the stationary microscopic trace that image formation spreads uniformly over the image \cite{marra2019gans, Corvi_2023_CVPR}. 

Crucially, \method is trained exclusively on real images. Generators leave fingerprints such as up-sampling artifacts and unnatural pixel-level correlations \cite{marra2019gans, Wang_2023_ICCV, Corvi_2023_CVPR, npr2024}, but these are generator-specific and shift with every new architecture. We showcase that the representation learned by the encoder is competitive in scenarios where no fake image is provided (i.e., zero-shot), where a few samples are provided (i.e., few-shot), and where fake samples from one generator are provided, fitting a linear classifier (i.e., linear probing). As can be seen in Fig.~\ref{fig:ft_overview}, we explore the potential of \method for \textit{i)} AI-generated image detection and \textit{ii)} unsupervised clustering of image sources. In summary, the main contributions are as follows.

\begin{itemize}
    \item \textbf{\method framework.} We propose a novel \ac{ssrl} framework that mitigates the bias towards semantic representations present in standard \ac{ssl} methods. We showcase that applying a residual extractor along with spatially disjoint crops favors the stationary micro-statistics of the image acquisition pipeline as the dominant signal the objective can exploit. Furthermore, \method is trained exclusively on real images, using a lightweight encoder with 3.0M parameters. We release the code, weights, and data splits for reproducible research.

    \item \textbf{Proficiency in Downstream Forensic Tasks.} We showcase the utility of \method in two Downstream Tasks: DT 1) Detection of AI-Generated Images, and DT 2) Clustering of Image Sources. For the detection task, a single GMM is fitted offline using only real image embeddings, and a real/AI-generated test image is flagged by its density under that model, achieving 97.21\% mean AUC across 26 generative architectures. Furthermore, we also assess the performance in few-shot and linear probing scenarios, where a limited number of fake images is provided. For image source clustering, we use $k$-Means and HDBSCAN clustering algorithms, fitted using the real/AI-generated image representations extracted from \method. Our proposed representations allow for clustering even when the number of clusters is not given.

\end{itemize}

\section{Related Work}
\label{sec:rel_work}

\noindent\textbf{AI-Generated Image Detection.} Generative pipelines leave subtle forensic traces, largely inherited from the architectural priors of decoders that map latent representations back to pixel space \cite{marra2019gans, Corvi_2023_CVPR, ganprintr}. Early detectors exploit these traces through supervised binary classification on a single generator family \cite{wang2020cnn}, sometimes under an explicit prior: LGrad \cite{lgrad2023} uses per-channel RGB gradients and NPR \cite{npr2024} targets the pixel correlations left by up-sampling. These models are strong in-domain but degrade sharply on unseen generators. Generalization has since been pursued by adapting vision-language backbones \cite{ojha2023towards}: EFFORT \cite{yan2024effort} and OmniAID \cite{guo2026omniaid} both steer CLIP-L/14 towards forensic patterns while preserving its semantic knowledge, the latter through a \ac{moe} decomposition. DIRE \cite{Wang_2023_ICCV} scores images by their diffusion reconstruction error before a binary classifier, while \citet{zhang2025detecting} completely dispense training, identifying an image by its distance from the natural-image manifold in the DINOv2 space. The latter shares our premise that real data alone suffices to detect fake images, yet builds it on a representation optimized for semantics. \method is trained so that the representations steer away from semantic content towards capturing the statistical fingerprint of image formation.

\vspace{4pt}
\noindent\textbf{Unsupervised Clustering of Image Sources.} Beyond binary detection, attributing a synthetic image to its source in an open-set scenario remains a formidable challenge. \citet{self-descriptors-cvpr} extract forensic self-descriptions for zero-shot detection and unsupervised clustering, but refine them per image through \ac{tto}, reporting approximately 0.11 images per second. 

\vspace{4pt}
\noindent\textbf{Self-Supervised Representation Learning.} Regularized-based \ac{ssl} \cite{balestriero2023cookbook, lecun2022path} learns from unlabeled corpora by making an encoder agree on two augmented views of an image. The different approaches differ mainly in how collapse is avoided: repelling negative pairs \cite{moco, mocov2}, matching a student to a momentum teacher \cite{dinov1, dinov2, dinov3}, or leveraging covariance regularized methods \cite{bt, bardes2022vicreg}. All rest on the same premise that the augmentations declare what the representation must discard and are chosen so that semantic content survives. SPAI \cite{spai2025} is the closest attempt to apply \ac{ssl} to a forensic task, pretraining with a masked spectral objective \cite{xie2022masked} before fitting a supervised binary head shaped by the generators seen in training, while NoisePrint++ \cite{guillaro2023trufor} used a supervised contrastive objective \cite{sup_contrastive_learning}, where they carefully selected positive and negative pairs considering different forensic attributes. \citet{self-consistency} proposed a self-consistency approach based on EXIF metadata to detect manipulated images. In contrast, our proposed \method architecture changes the input signal from standard RGB images to residuals along with spatially disjoint views, steering the optimization towards forensic representations and away from semantics without the need for contrastive negative samples. 

\section{Proposed Method: \method}
\label{sec:method}

We introduce \method, a novel \ac{ssrl} framework designed to extract forensic representations without labels of the image generation process using only a corpus of real images. In this section, we formulate the underlying objective, detail our feed-forward architecture, and explain how strict spatial and sampling constraints enforce the decoupling of semantics from physical fingerprints.

\subsection{Background}
\label{subsec:background}

\noindent\textbf{Forensic Residuals.} Any digital image $I \in \mathbb{R}^{H \times W}$ can be conceptualized as a combination of its macroscopic semantic content $C$ and its microscopic high-frequency noise fingerprint $\Psi$, e.g. color filter arrays, camera sensor Photo-Response Non-Uniformity (PRNU), or generative up-sampling/quantized artifacts. An example of this is SRM filters \cite{srm-filters}, which act as a set of different linear high-pass filters that aim to isolate $\Psi$. \citet{self-descriptors-cvpr} proposed to model an image as an additive operation between both terms $I(x,y) = C(x,y) + \Psi(x, y)$, where $x$ and $y$ are the 2D pixel coordinates. To isolate $\Psi$, they introduced a constrained learnable filter $\mathcal{K}$, optimized to estimate the semantic content $\hat{C}$, thus extracting the residual by subtracting the estimation $\hat{C}$ from the original image $I$. The resulting tensor $I_{res} = \mathcal{F}(I) \in \mathbb{R}^{H \times W \times 8}$ theoretically suppresses low-frequency semantics while amplifying $\Psi$. However, even with these elaborated residual extractors, macroscopic edges often survive this filtering, which may lead to semantic entanglement when processed by standard deep architectures.

\vspace{4pt}
\noindent\textbf{Barlow Twins Objective.} To learn meaningful representations without labels, we build upon the Barlow Twins objective \cite{bt}. Given two views of an image $I$, a neural network composed of an encoder $f_{\theta}$ and a projector $g_{\phi}$ produces the embeddings $E_1, E_2 \in \mathbb{R}^{B \times D}$ (batch size $B$ and embedding size $D$). The embeddings are batch-normalized, and each element $\mathcal{C}_{ij}$ of the cross-correlation matrix $\mathcal{C} \in \mathbb{R}^{D \times D}$ across a batch of views is calculated as:

\begin{equation}
    \mathcal{C}_{ij} = \frac{1}{B} \sum_{b=1}^{B} E_{1,b,i} E_{2,b,j}
    \label{eq:corr_matrix}
\end{equation}

\noindent where $i$ and $j$ correspond to the index along the activations in the embedding layer throughout the batch $B$. The loss function minimizes the difference between $\mathcal{C}$ and the identity matrix $\mathcal{I}$:

\begin{equation}
    \mathcal{L}_{\mathcal{BT}} = \sum_{i} (1 - \mathcal{C}_{ii})^2 + \lambda \sum_{i} \sum_{j \neq i} \mathcal{C}_{ij}^2  
    \label{eq:bt_loss}
\end{equation}

By driving the diagonal elements to $1$ (invariance) and off-diagonal elements to $0$ (redundancy reduction), the network is forced to find the underlying invariant factors of variation between the two views, without the need for negative samples.

\subsection{Forensic Twins: A Feed-Forward Architecture for Semantic Decoupling}
\label{subsec:architecture}

Fig.~\ref{fig:ft_overview} provides an overview of our proposed \method. Given a batch of color images $X_{RGB} = \{I_i^{RGB}\}_{i=1}^N$, where $I_i^{RGB} \in \mathbb{R}^{H \times W \times 3}$, we first apply a non-invasive transformation pipeline that preserves as much forensic information from the pristine pixel distribution as possible. This deliberately excludes transformations that alter the forensic micro-structures of the image. Therefore, we define the collection of valid transformations as \textit{i)} random orthogonal rotations (i.e., 90º, 180º and 270º), and \textit{ii)} random flipping in the vertical and horizontal axes. Both transformations are drawn once per image and shared by the two views, so that the location from which each view is cut remains the only factor distinguishing them.

The transformed color images are then converted to grayscale, denoted as $X_{g} = \{I_i^{g}\}_{i=1}^N$, where $I_i^{g} \in \mathbb{R}^{H \times W}$, and normalized to be in the $[0,1]$ range by dividing each pixel value by 255. The semantic content $C$ is then severely suppressed by applying a \ac{fre} $\mathcal{F}$ from \cite{self-descriptors-cvpr}, which is frozen, resulting in $X_{res} = \{I_i^{res}\}_{i=1}^N$, where $I_i^{res} \in \mathbb{R}^{H \times W \times 8}$, where 8 is the number of channels of the \ac{fre}. These residual images contain multiple complementary channels that capture different patterns of residual high-frequency information \cite{self-descriptors-cvpr}. However, residual images $X_{res}$ inherently retain structural semantic edges that can be leveraged by the encoder to create an optimization shortcut, steering the representations toward semantics rather than the underlying forensics. 

To ensure that the invariant factor maximized by $\mathcal{L}_{\mathcal{BT}}$ is this micro-statistical noise $\Psi$ and not the semantic content $C$ left by \ac{fre} \cite{self-descriptors-cvpr}, we propose to avoid standard overlapping views in favor of spatially disjoint ones. The statistical fingerprint left by physical acquisition devices in real images and generative artifacts in AI-generated images is correlated, as stationary patterns occur throughout the image plane \cite{marra2019gans}. This means that spatially disjoint views carry cross-correlated fingerprints from the image while restricting semantic information. Given an arbitrary residual image from the batch $I_i^{res}$, we sample two spatial coordinates $(x_1, y_1)$ and $(x_2, y_2)$ to extract two disjoint crops \cite{chai2020makes}, $R_1 \in \mathbb{R}^{H_1 \times W_1 \times 8}$, and $R_2 \in \mathbb{R}^{H_2 \times W_2 \times 8}$, which may or may not have different resolutions, and use them as input views for the encoder $f_{\theta}$.

Both batches of disjoint views $R_1$ and $R_2$ are then fed into our representation learning pipeline, composed of a feature encoder $f_{\theta}$ and a non-linear projector $g_{\phi}$. For batched tensor operations, all $R_1$ crops within a mini-batch share a common spatial size $H_1 \times W_1$, and independently all $R_2$ crops share a common size $H_2 \times W_2$. However, $H_1 \times W_1$ need not equal $H_2 \times W_2$. Both crops sizes are arbitrarily re-sampled independently ---across fixed bucket sizes of 96, 128, 192, 224, 256, and 384--- and at every training iteration, which motivates using 2D rotary positional embeddings \cite{rope, dinov3} in the encoder $f_\theta$, so that a single set of weights handles crops of arbitrary size for both views. The representation produced by the encoder is indicated as $Z_{rep}^{(v)}$, where $v$ denotes the view (i.e., $R_1$ or $R_2$) from which the representation is extracted. Finally, to compute the self-supervised objective, these representations are mapped through the projector network $g_{\phi}$ to obtain the embeddings $E_v = g_{\phi}(Z_{rep}^{(v)})$. The cross-correlation matrix $\mathcal{C}$ is then constructed from $E_1$ and $E_2$. Since the underlying micro-statistical noise $\Psi$ remains the dominant shared signal between the disjoint views, the Barlow Twins loss $\mathcal{L}_{\mathcal{BT}}$ is coerced to align forensic cues rather than semantics.

\subsection{Downstream Tasks (DT): Forensic Applications}
\label{subsec:downstream}

Following standard practices in self-supervised learning \cite{bt}, once \method is pre-trained exclusively on real images, the non-linear projector $g_{\phi}$ is discarded. The frozen encoder $f_{\theta}$ is retained to extract robust micro-statistical representations $Z_{rep}$ directly from the summary representation (i.e., the \texttt{[CLS]} token in the \ac{vit}). These embeddings can then be leveraged for various downstream forensic tasks without requiring any fine-tuning or test-time optimization.

\vspace{4pt}
\noindent\textbf{DT 1: AI-Generated Image Detection.} 
We explore three different scenarios for AI-generated image detection: zero-shot, few-shot, and linear probing. For the zero-shot scenario, we approach AI-generated image detection as an anomaly detection problem in the latent space. We fit a GMM to the real image distribution, estimating the density $P(Z)$, where $Z$ are the embeddings of real images. Given a test image, we compute its embedding $Z_{test}$ and score it as the negative log-likelihood $\mathcal{S}_{anomaly}(Z_{test}) = - \log P(Z_{test})$. For the few-shot scenario, we take 25 real images and 25 fake images from a single generator (e.g., SD1.4) and train a $k$-NN classifier. In the linear probing scenario, we train a linear classifier with 1000 real images and 1000 images from a single AI-generator (e.g., SD1.4).

\vspace{4pt}
\noindent\textbf{DT 2: Clustering of Image Sources.}
Beyond binary detection, the structural topology of the learned latent space allows clustering distinct Generative AI architectures based on their unique up-sampling or decoding artifacts. For source attribution, we extract representations $Z_{rep}$ from an unlabeled mixture of real images and various AI-generated images. To perform unsupervised clustering of real/AI-generated image sources, we apply two different algorithms \textit{i)} the $k$-Means algorithm \cite{kmeans} directly on the encoder's output, where the number of sources is known, and \textit{ii)} the HDBSCAN algorithm \cite{campello2013density}, where no prior information is given regarding the number of clusters. 
\section{Experimental Results}
\label{sec:results}

\noindent\textbf{Optimization Details.}
\method optimization was conducted on a local compute node equipped with four NVIDIA RTX 4090 GPUs. \method was trained for 80 epochs using AdamW, with a base learning rate of 0.0015 and a weight decay of 0.0025. The redundancy-reduction term from Eq.~\ref{eq:bt_loss} is set to $\lambda=5e-5$. We employ a learning rate schedule consisting of a 15-epoch linear warm-up, followed by a cosine annealing decay down to $0.000001$. Given the VRAM constraints of the GPUs, we leveraged \texttt{bfloat16} format and FlashAttentionV2 \cite{flash-attn-2} for optimized memory consumption, with a batch size of 320. The output dimension of the projector $g_{\phi}$ is set to 8,192 across all experiments. Regarding the training of the GMM, we use the same real images as the encoder pre-training, with a number of components experimentally set to $K=1$ using the diagonal covariance matrix. We perform ablation studies regarding the GMM in Sec. \ref{ssec:ablations}. Regarding view size, we considered different bucket sizes: 96, 128, 192, 224, 256, and 384, which depend on each image original size. These buckets are assigned to each image, depending on the size that fits two disjoint crops in the same image.

\begin{table*}
\centering
\resizebox{\textwidth}{!}{%
\begin{tabular}{@{}l c r r cccc@{}}
\toprule
\textbf{Method} & \textbf{Real Only?} & \textbf{\#Params} & \textbf{Lat. (ms)} & \textbf{GANs (8)} & \textbf{Diffusion (13)} & \textbf{Commercial (5)} & \textbf{Global (26)} \\
\midrule
CNNDet\textsubscript{\textcolor{gray}{[CVPR20]}}  & \xmark & 23.5M  &    \underline{2.94} & 91.83 & 68.19 & 62.83 & 74.43 \\
UFD\textsubscript{\textcolor{gray}{[CVPR23]}}     & \xmark & 304.0M &    8.14 & 97.73 & 82.69 & 85.03 & 87.77 \\
LGrad\textsubscript{\textcolor{gray}{[CVPR23]}}   & \xmark & 46.6M  &   12.50 & 91.32 & 77.02 & 67.61 & 79.61 \\
NPR\textsubscript{\textcolor{gray}{[CVPR24]}}     & \xmark & \underline{1.4M}   &    \underline{1.61} & 90.55 & 89.45 & 68.81 & 85.82 \\
SPAI\textsubscript{\textcolor{gray}{[CVPR25]}}    & \xmark & 139.9M &   26.42 & 93.08 & 92.45 & 91.38 & 92.44 \\
EFFORT\textsubscript{\textcolor{gray}{[ICML25]}}  & \xmark & 504.6M &   10.21 & 98.31 & \underline{97.01} & \underline{96.79} & \underline{97.37} \\
OmniAID\textsubscript{\textcolor{gray}{[ICLR26]}} & \xmark & 508.8M &   40.64 & \underline{99.40} & \underline{97.66} & 92.30 & \underline{97.17} \\
ConV\textsubscript{\textcolor{gray}{[NeurIPS25]}} & \cmark & 304.4M &   16.82 & 73.95 & 69.02 & 69.77 & 70.68 \\
FSD\textsubscript{\textcolor{gray}{[CVPR25]}}     & \cmark & \underline{6.2M}   & 2,413.66 & 57.69 & 52.53 & 43.84 & 52.45 \\
\midrule
Ours (zero-shot, GMM) & \cmark & \underline{3.0M} & 5.73 & 97.81 & \underline{97.33} & \underline{95.92} & \underline{97.21} \\
Ours (few-shot, $k$-NN) & \xmark & \underline{3.0M} & \underline{5.37} & \underline{98.82} & 92.89 & \underline{96.47} & 95.40 \\
Ours (linear probe) & \xmark & \underline{3.0M} & 5.60 & \underline{98.79} & 93.07 & 94.45 & 95.10 \\
\bottomrule
\end{tabular}%
}
\caption{AI-generated image detection performance (AUC \%), averaged per category across 26 generative architectures and 2 real-image distributions (ImageNet-1k, MS-COCO). Parameter counts and latency are reproduced from the full cost breakdown in Appendix~\ref{app:complexity} (for \method we include \ac{fre}'s module from \cite{self-descriptors-cvpr} in the computational cost and complexity statistics); latency is a median over 500 ImageNet-1k images on a single RTX 4090. Regarding the results, we indicate whether per-generator family average and the global average, as well as if the method is trained using only real images. Top-3 methods are \underline{underlined}.} 
\label{tab:det_results_full}
\vspace{-12pt}
\end{table*}

\vspace{4pt}
\noindent\textbf{Training Dataset.} Following recent approaches in the field of AI-generated image detection \cite{spai2025, self-descriptors-cvpr, zhang2025detecting, Corvi_2023_CVPR}, our proposed \method is trained using only real images from ImageNet-1k \cite{imagenet} and MS-COCO \cite{ms-coco}. Specifically, 250,000 images are gathered from the ImageNet-1k training set and 241,690 images from MS-COCO training data, resulting in 491,690 images in total. In the few-shot variant, we leverage 25 real images from MS-COCO and ImageNet-1k training sets and 25 fake images from SD1.4 generator. For the linear probe variant, we increase the number of real and fake images to 1000 each. In the AI-image-generated task validation set, we used 750 real images from the MS-COCO and ImageNet-1k validation splits and SD1.4 \cite{Rombach_2022_CVPR} as the AI-generated source from \cite{zhu2023genimage}, which is excluded from the evaluation set.

\subsection{AI-Generated Image Detection}

\noindent\textbf{Experimental Set-Up.} We evaluate \method against state-of-the-art AI-generated image detection methodologies. This involves the following methods: CNNDet~\cite{wang2020cnn}, LGrad~\cite{lgrad2023}, UFD~\cite{ojha2023towards}, NPR~\cite{npr2024}, ConV~\cite{zhang2025detecting}, FSD~\cite{self-descriptors-cvpr}, SPAI~\cite{spai2025}, EFFORT~\cite{yan2024effort}, and OmniAID \cite{guo2026omniaid}. As in \cite{self-descriptors-cvpr}, we use AI-generated images from different well-known datasets: OSSIA~\cite{ossia-dataset}, DMID~\cite{dmid-dataset}, and SynthBuster~\cite{synthbuster-dataset}. Some notable generators are: StyleGAN family \cite{sauer2022stylegan, alaluf2022third}, ProGAN \cite{karras2018progressive}, EG3D \cite{chan2022efficient}, GLIDE \cite{glide}, Stable Diffusion family \cite{ esser2024scaling, podell2024sdxl}, and DiT \cite{peebles2023scalable}. A complete breakdown of the different generators considered in the evaluation is detailed in the Appendix \ref{app:detailed_generators}. Furthermore, we add AI-generated images from commercial generators (i.e., Firefly \cite{adobe2024firefly}, Midjourney v6 \cite{midjourney2024v6}, and Flux \cite{blackforestlabs2024flux}) extracted from online resources\footnote{\url{https://www.kaggle.com/datasets/aryankaushik005/firefly}}$^{,}$\footnote{\url{https://huggingface.co/datasets/brivangl/midjourney-v6-llava}}$^{,}$\footnote{\url{https://huggingface.co/datasets/Rapidata/Flux_SD3_MJ_Dalle_Human_Alignment_Dataset}} (e.g., Kaggle and HuggingFace), evaluating both \method and previous state-of-the-art methods in 26 total AI generators, which we divide into three fundamental subsets: GAN-based generators, Diffusion-based generators, and Commercial generators. For real images, following previous approaches in the literature \cite{self-descriptors-cvpr, Corvi_2023_CVPR, zhu2023genimage}, 750 images are selected from the held-out test split of ImageNet's and MS-COCO, disjoint from the 491,690 images used for pre-training. For AI-generated images, a total of 750 images per generator were considered across the datasets (i.e., OSSIA, DMID, SynthBuster), balancing across each real vs. generator evaluation. We observed evidence in these benchmarks of a confound in terms of image resolution (see Appendix~\ref{app:resolution}). To avoid it, we take a $256 \times 256$ center crop, which is the most common resolution across all fake images.

\vspace{4pt}
\noindent\textbf{Results.} Table~\ref{tab:det_results_full} reports the performance. Our proposed method achieves competitive results on average (97.21\% AUC) in the zero-shot scenario, ranking second within 0.2 points with EFFORT and OmniAID. A simple ranking over the global average column might hide the important advantages of \method. First, seven of the nine state-of-the-art methods are calibrated on AI-generated images (see the \textit{Real-Only?} column), and the cost of a decision (latency) spans three orders of magnitude. Considering similar state-of-the-art methods that never observe AI-generated images (i.e., ConV and FSD), \method outperforms ConV by 26.5 points (70.68 vs. 97.21\% AUC) with a backbone that is two orders of magnitude smaller (3.0M vs. 304.4M parameters). FSD performs near random with 52.45\% AUC (see Appendix~\ref{app:resolution}), while taking 2,413.66 ms per decision against our ~5.5 ms. The few-shot and linear probe variants perform worse than the zero-shot counterpart but remain competitive overall, with 95.40\% and 95.10\% AUC, respectively.

The per-category performance columns also provide interesting insights. In a zero-shot scenario, it achieves its lowest performance on commercial generators (95.92\%), although 3.62\% above OmniAID, which falls from 99.40\% on GANs to 92.30\% there. The few-shot variant improves in GANs and Commercial generators, with 98.82\% and 96.47\% but under-performs in diffusion models (92.89\%). The linear probe excels at detecting GAN based models with 98.79\%, but its performance decreases the most in diffusion (93.07\%). Other methods trained using AI-generated images degrade far more sharply: NPR drops from 90.55\% on GANs to 68.81\% on commercial ones, and CNNDet, LGrad and UFD span 29.00\%, 23.71\%, and 15.04\% between their best and worst category. 

To summarize, two methods rank on par with us on the global average, EFFORT and OmniAID, both with a nearly 500M-parameter backbone trained with supervision on AI-generated images. Our claim is that the proposed representation learned from real images alone comes very close to the strongest supervised detector, with only 3.0M parameters and a single forward pass in inference (around 5.5 \,ms). 

\paragraph{Computational Cost.} The columns \#Params and Lat. (ms) of Table~\ref{tab:det_results_full} report only parameters and latency. For our proposed method, we consider the \ac{fre} in the total cost. \method is the cheapest entry in the benchmark on memory 516 MB, and the second least parameterized behind NPR, while on par with CNNDet in terms of GFLOPs (10.7). NPR and CNNDet are faster, at 1.61 ms and 2.94 ms, as their architectures do not rely on attention based models (i.e., ViT), in contrast to \method. FSD's per-image solve costs 2537.0 GFLOPs and 11342 MB. Additional details on the computational advantages of \method are included in Appendix~\ref{app:complexity}.

\subsection{Unsupervised Clustering of Image Sources}
\label{sec:clust_img_sources}

\vspace{4pt}
\noindent\textbf{Experimental Set-Up.} We also evaluate the capabilities of \method for image source clustering against similar self-supervised methods that only consider real images for training: FSD~\cite{self-descriptors-cvpr} and ConV~\cite{zhang2025detecting}, as well as methods that are optimized by seeing fake samples: NPR~\cite{npr2024}, OmniAID~\cite{guo2026omniaid} and EFFORT~\cite{yan2024effort}. Furthermore, we include CLIP's vision backbone, given its many uses in AI-generated supervised detection methods \cite{ojha2023towards, cozzolino2024raising, yan2024effort, guo2026omniaid}. We follow the unsupervised clustering protocol from \cite{self-descriptors-cvpr} and use real images from ImageNet-1k test set ---which is disjoint from the training set used to train the encoder--- and AI-generated images from ProGAN \cite{karras2018progressive}, StyleGAN3~\cite{alaluf2022third}, GLIDE~\cite{glide}, SD3~\cite{esser2024scaling}, DALLE-3~\cite{betker2023improving}, Midjourney v6, and Firefly. We explore $k$-Means \cite{kmeans} and HDBSCAN \cite{campello2013density} as different clustering methods to assess the structure of the latent spaces of the competing methods and its grouping properties when \textit{i)} the number of sources is given (i.e., $k$-Means) and \textit{ii)} no prior information about clusters is given (i.e., HDBSCAN). 

\begin{table}
    \begin{tabular}{@{}lccc@{}}
        \toprule
        \textbf{Method} & \textbf{Real Only?} & \textbf{\textit{$k$-Means}} & \textbf{HDBSCAN} \\
        \cmidrule(lr){1-4}
        NPR & \xmark & 45.6 / 44.3 & 12.6 / 4.5 \\
        OmniAID & \xmark & 63.6 / 60.2 & 14.1 / 38.8 \\
        EFFORT & \xmark & \underline{70.7 / 73.5} & 17.2 / 40.0 \\
        \midrule
        CLIP-L/14 & \cmark & 58.2 / 51.1 & 12.0 / 32.9 \\
        FSD & \cmark & 29.7 / 12.7 &  7.2 / 7.8  \\
        ConV & \cmark & 35.6 / 31.5 & 12.0 / 31.1 \\
        \midrule
        Ours & \cmark & 62.7 / 70.3 & \underline{50.1 / 66.6} \\
        \bottomrule
    \end{tabular}
    \caption{Comparison for unsupervised clustering of image sources in terms of accuracy (Acc.) and Normalized Mutual Information (NMI). The number of sources is 8. Best result is \underline{underlined}.}
    \label{tab:cluster_results}
    \vspace{-14pt}
\end{table}

\vspace{4pt}
\noindent\textbf{Results.} Table~\ref{tab:cluster_results} reports unsupervised clustering of image sources using $k$-Means and HDBSCAN in terms of accuracy (Acc.) and Normalized Mutual Information (NMI).

When the number of clusters is known (i.e., $k$-Means), we observe that EFFORT clearly clusters the generators more proficient overall from the methods that have seen fakes in the training stage, with 70.7\% of accuracy. For methods that never saw fake images in training, \method is the best one, with 62.7\% accuracy, followed by CLIP-L/14 with 58.2\%. ConV leverages DINOv2 representations which seem to be not enough to properly identify sources (35.6\%). We attribute this mainly because DINOv2 is oriented to extract semantic representations (see Sec.~\ref{sec:forensic_representation_analysis}).

Intriguingly, we observe that when the number of clusters is not given (i.e., HDBSCAN), \method is the best performing, with 50.1\% in terms of accuracy, 32.9 accuracy points above the second best. Other powerful methods such as EFFORT (17.2\%) or OmniAID (14.1\%) do not provide representations of clustering that can be performed when no prior information is given on the number of clusters. This indicates that sources form dense, separated regions in \method's latent space, recoverable without knowing their number.

\subsection{Ablation Studies}
\label{ssec:ablations}

Our ablation studies vary different aspects of \method starting from a baseline ViT-Tiny (six blocks, patch-8 with a linear stem, \texttt{[CLS]} token and an embedding dimension of 192) on RGB data with $\lambda=0.005$, trained for 80 epochs (batch size 320) on the 491,690 ImageNet and MS-COCO images. Views are disjoint crops of the training images, and non-destructive augmentations are applied (i.e., flips and orthogonal rotations). For each ablation, we report the AUC (\%) in the detection task over the validation set in the most-challenging scenario, zero-shot detection.

\vspace{4pt}
\noindent\textbf{Residual Extractor.} When using raw RGB images as input, the baseline aligns views through semantics and barely detects any AI-generated image (69.18\% AUC). Placing a residual extractor before the encoder shifts what it learns: fixed SRM filters \cite{srm-filters} recover most of the gap (95.40\%), and the learned \ac{fre} from \cite{self-descriptors-cvpr} adds a further 1.59 points (96.99\%). Suppressing semantic content, rather than the specific filter, is what enables detection, with the \ac{fre} as its strongest instance. We use the \ac{fre} for the rest of the ablations as an instance of residual extraction since it is a key component of our architecture.

\vspace{4pt}
\noindent\textbf{GMM.} We consider four different numbers of components for the GMM, these are $K \in [1, 2, 4, 8]$. In addition, we considered two scenarios: \textit{i)} full covariance matrix and \textit{ii)} only diagonal matrix after applying a PCA, retaining 99\% of the variance. The results are presented in Table \ref{tab:gmm}, where the diagonal covariance matrix configuration performed the best, particularly $K=1$ with an AUC of 98.78\%.

\begin{table}
    \centering
    \begin{tabular}{lcccc}
    \toprule
     \textbf{Cov. Matrix} & \textbf{\textit{K}=1} & \textbf{\textit{K}=2} & \textbf{\textit{K}=4} & \textbf{\textit{K}=8} \\
    \midrule            
    Full  & 97.83 & 97.58 & 96.80 & 97.00 \\
    Diagonal & \underline{98.78} & 98.70 & 98.62 & 98.72 \\
    \bottomrule
    \end{tabular}
    \caption{Validation performance in terms of AUC(\%) using different configurations of the GMM regarding the number of components $K$ and the covariance matrix. Best result is \underline{underlined}.}
    \label{tab:gmm}
    \vspace{-12pt}
\end{table}

\begin{figure}
    \centering
    \includegraphics[width=0.85\linewidth]{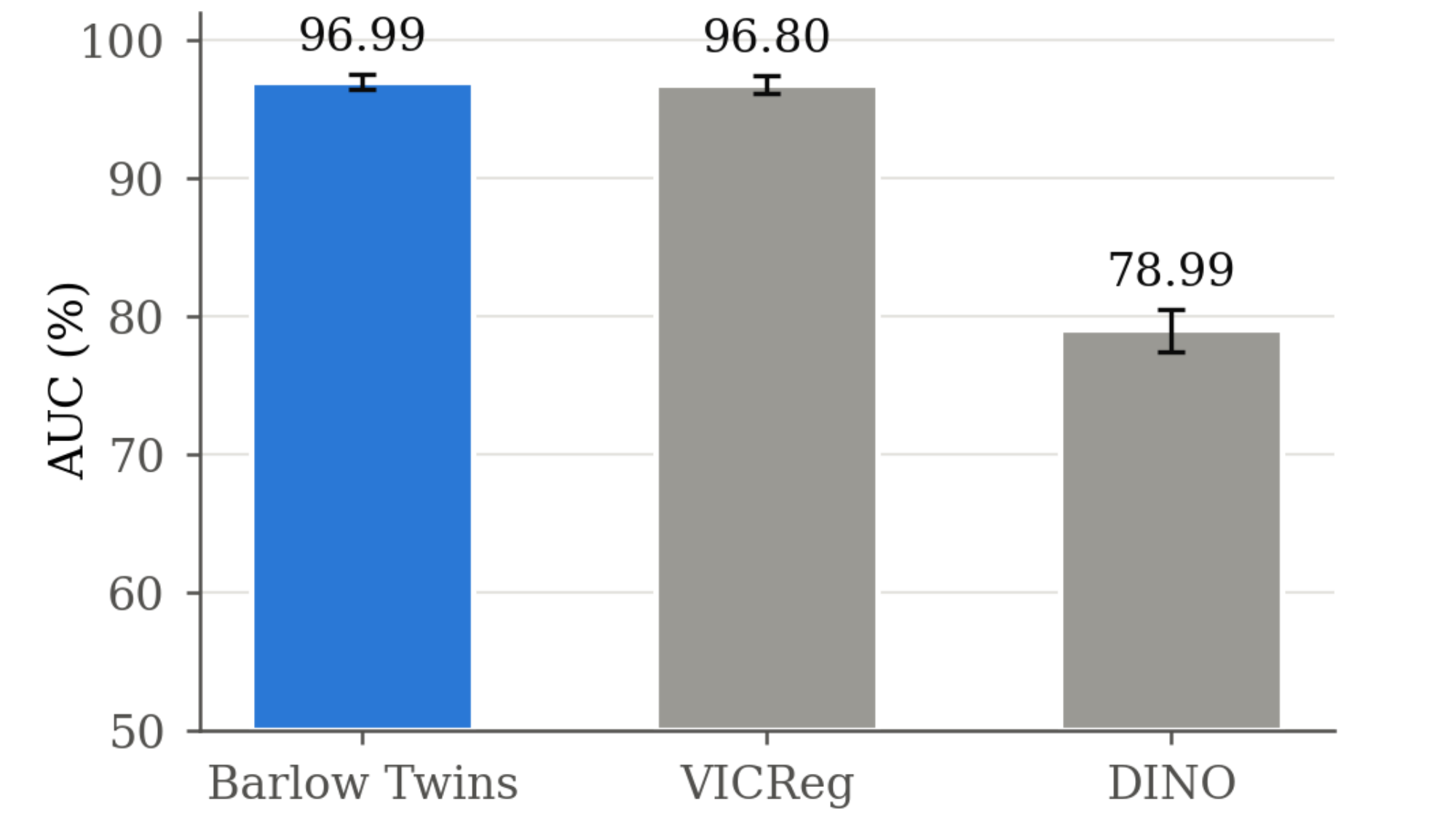}
    \caption{Validation performance in terms of AUC(\%) of different self-supervised methods, with their default hyper-parameters.}
    \label{fig:objectives}
    \vspace{-12pt}
\end{figure}

\vspace{4pt}
\noindent\textbf{SSL Objectives.}
We inspect the effect of other self-supervised objectives in our proposed framework. Beyond Barlow Twins' objective, we also consider VICReg~\cite{bardes2022vicreg} as another covariance-regularized objective. We also consider DINO~\cite{dinov1}, since it is a widely established self-distillation method for self-supervised learning. Fig.~\ref{fig:objectives} shows that both covariance regularized methods perform practically on par, with Barlow Twins slightly ahead (96.99\% AUC), however, the self-distillation method, DINO, fails to perform accordingly to the detection task, achieving only 78.99\% AUC. We partly attribute this to self-distillation dependence on careful hyper-parameter selection and multi-view distortions \cite{dinov1, byol}, which our framework does not contemplate. 

\vspace{4pt}
\noindent\textbf{Augmentations.} We ablate how the two views (i.e., $R_1$ and $R_2$) are built. Overlapping instead of disjoint crops reduces AUC to 90.19\% AUC, -6.80 points compared to the disjoint strategy, as shared pixels likely let the encoder match views through content rather than forensic traces. DINOv2-style \cite{dinov2} destructive augmentations are far more harmful (57.76\% AUC, decreasing performance 39.23 points): they overwrite the residual patterns both views share, and training fails to align them.

\vspace{4pt}
\noindent\textbf{Regularization.} We explore different values regarding the redundancy term $\lambda$ in Barlow Twins' objective (Eq.~\ref{eq:bt_loss}), where we consider the following three different values $\lambda \in $ [2e-4, 5e-5, 1e-5], from the default 5e-3. We observe that this value grants small performance improvements in terms of AUC (\%), as 2e-4 enables an increase of 96.99\% $\rightarrow$ 97.41\%, and 5e-5 increases this slightly further, achieving the best validation AUC with 98.78\%. Finally, 1e-5 reduces performance compared to 5e-5, with 98.73\% $\rightarrow$ 98.13\%.
\begin{figure*}
    \centering
    \includegraphics[width=\textwidth]{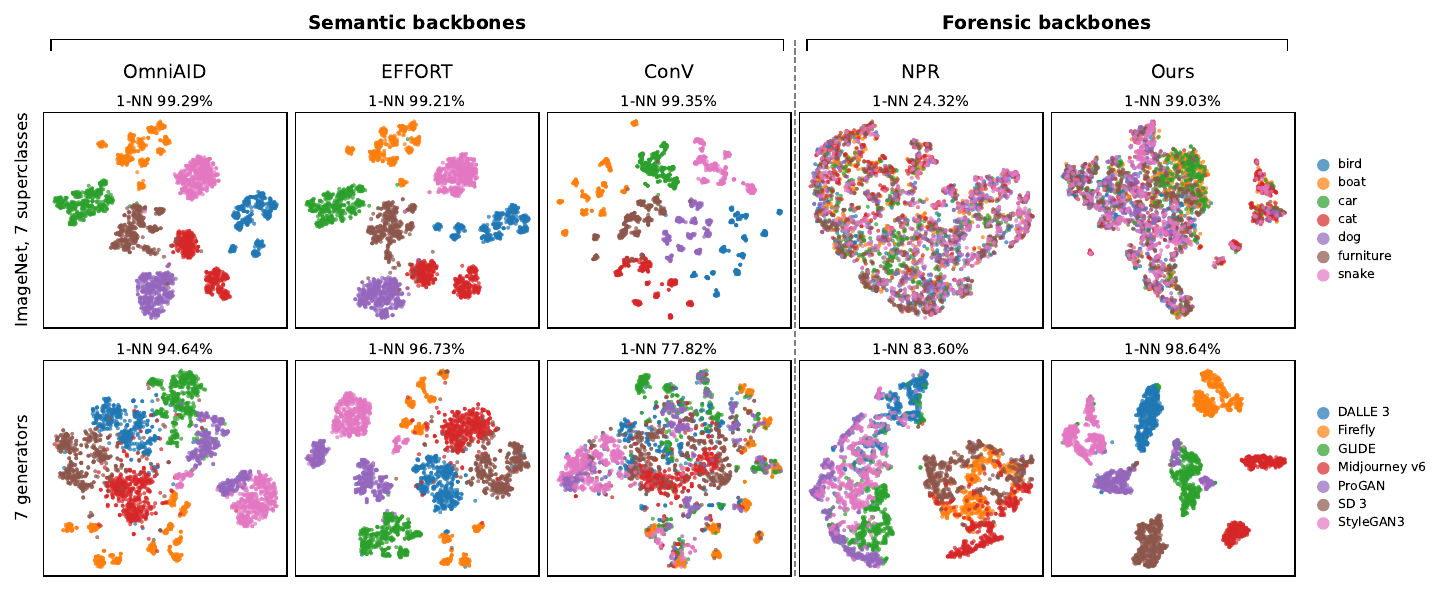}
    \vspace{-22pt}
    \caption{t-SNE comparison between 7 semantic classes from the ImageNet-1k dataset (top) and images from different AI source generators, across GANs, Diffusion Models and Commercial Models (bottom) from \method and different methods considered in Sec. \ref{sec:clust_img_sources}.}
    \label{fig:tsne}
    \vspace{-15pt}
\end{figure*}

\section{Forensic Representation Analysis}
\label{sec:forensic_representation_analysis}
We next analyze the representations of methods that adapt a general Vision Foundation Model (VFM) (i.e., CLIP or DINO) to forensic tasks and methods built by design on forensic information. A representation can separate image sources for two very different reasons. It may encode the low-level traces each generation pipeline leaves in the pixels, which is what a forensic detector should capture. Or it may encode \emph{what} the images depict: generators also differ in their content distribution, so a purely semantic embedding can attribute sources through content alone. Source clustering cannot tell these apart. To disentangle them, we measure every representation with the same probe on a task solvable \emph{only} through content and AI source classification.

\vspace{4pt}
\noindent\textbf{Experimental Set-Up.} For the semantic task we take 3{,}360 images from seven ImageNet-1k superclasses (bird, boat, car, cat, dog, furniture, and snake). All images come from the validation split, so none of them was seen during the pre-training of our encoder, which only uses the training split. For the source task, we use seven generators spanning the three families of the benchmark: ProGAN and StyleGAN3 (GANs); GLIDE and Stable Diffusion 3 (Diffusion); and Firefly, DALLE~3, and Midjourney v6 (Commercial), with 480 held-out test images per generator. Therefore, both tasks have seven balanced classes and a chance level of $1/7 = 14.29\%$. The considered methods are ConV~\cite{zhang2025detecting}, EFFORT~\cite{yan2024effort} and OmniAID~\cite{guo2026omniaid} for adapted VFM and NPR~\cite{npr2024} and \method for forensic-based methods. All embeddings are $L_2$-normalized, and a $k$-NN ($k=1$) classifier with cosine distance is fitted to half the images of each class and evaluated on the other half.

\vspace{4pt}
\noindent\textbf{Results.} Fig. \ref{fig:tsne} shows the results of this experiment, where in the upper row we see the t-SNE of the semantic space across the considered models, and the $k$-NN accuracy performance. In the bottom row, we see an analogous visualization of the t-SNE space but for AI source generator. It is clear that semantics do not dominate the representations of NPR and \method. Especially, NPR shows a performance that is near random in terms of semantic accuracy, with 24.32\%, while \method seems to be more sensitive to semantic features (39.03\%). However, VFM-adapted methods appear to retain a large portion of semantic information, as all of them achieve ~99\% accuracy. However, for the AI source generator task, we observe that \method is capable of accurately identifying the source (98.64\%), with EFFORT (96.73\%) and OmniAID (94.64\%) behind. These results suggest that its attribution is primarily driven by forensic traces rather than content. These observations indicate that there is a semantic influence in the adapted VFM methods that allow them to be competitive in this task, which does not appear to apply to \method. We wonder whether, as the realism of AI-generated images inevitably increases, semantically influenced representations will remain competitive in forensic-related tasks.

\section{Conclusion}
\label{sec:conclusion}

This paper has presented \method, a self-supervised residual framework that learns the statistical fingerprint of real image formation without ever observing an AI-generated image. The design principle relies on preventing the encoder from taking the easy path towards semantic content by \textit{i)} applying a residual filter that suppresses the majority of semantic content from a given image, \textit{ii)} extracting disjoint crops to alleviate potential semantic shortcuts, as statistical fingerprints occur along the whole image plane, and \textit{iii)} removing destructive augmentation pipelines present in standard \ac{ssl} pipelines, as they remove most of the forensic signal from real images. 

We have tested the learned representations in two challenging tasks: 1) Detection of AI-generated Images, evaluating the performance across 26 different Generative AI architecture across different architectures (i.e., GANs, Diffusion Models, and Commercial); and 2) Clustering of Image Sources, where \method is capable of attributing image sources without \textit{i)} ever having seen any AI-generated image, and \textit{ii)} with a substantially weaker dependence on semantic content than VFM-based alternatives. 

Future work will address \method' limitations as improving performance on commercial architectures, as \method perform the worst in the zero-shot variant compared to GAN and Diffusion architectures.  Designing an objective that leverages dense encoder features to expand encoder functionalities to manipulation localization is also at the top of our priorities.
\section*{LLM Usage Statement}

The authors assume full responsibility for all content in this submission, regardless of any LLM or AI-tool use. LLM-based tools were used only in an assistive capacity, to edit or refine content produced by the authors, in the following areas: writing and editing of the text, related work and citations, code and implementation, design and analysis of the experiments, figures, and the presentation of research ideas and claims. No content was generated by these tools and no task was carried out autonomously by them. No LLM or AI tool was used for data creation, synthesis, or annotation. In addition, the full manuscript was provided to an LLM (Claude Anthropic) to obtain reviewer-style feedback.
\section*{Acknowledgements}

This project has been supported by PowerAI+ (SI4/PJI/2024- 00062 Comunidad de Madrid and UAM), Cátedra ENIA UAM-Veridas en IA Responsable (NextGenerationEU PRTR TSI-100927-2023-2), and TRUST-ID (PID2026-185599OB-I00 funded by MICIU/AEI/10.13039/501100011033).
{
    \small
    \bibliographystyle{ieeenat_fullname}
    \bibliography{main}
}

\clearpage
\setcounter{page}{1}
\maketitlesupplementary

\section{Image Resolution as a Shortcut}
\label{app:resolution}

\begin{figure*}
    \centering
    \includegraphics[width=\textwidth]{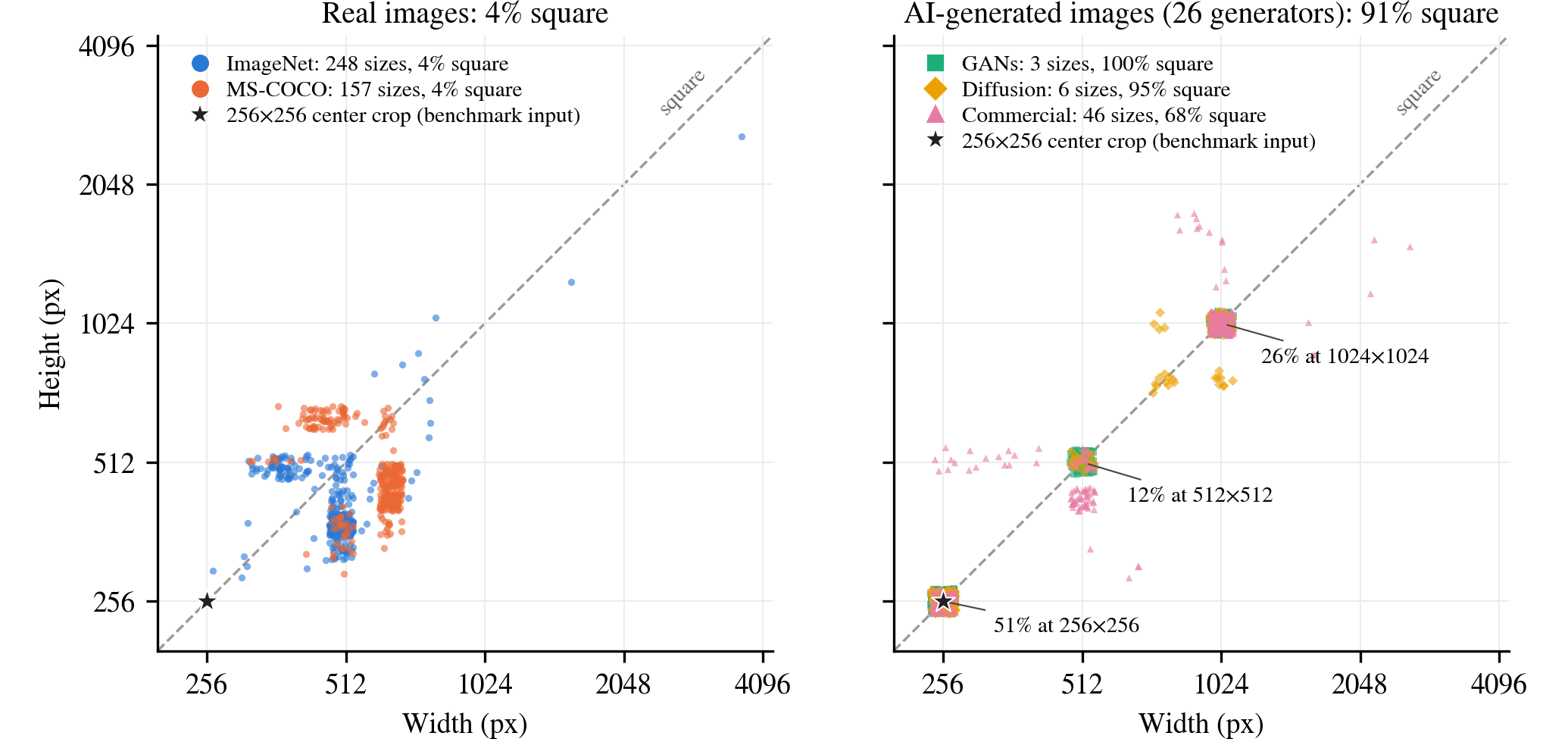}
    \caption{Height and width distributions of the benchmark's images in terms of pixels from OSSIA~\cite{ossia-dataset}, SynthBuster~\cite{synthbuster-dataset} and DMID~\cite{dmid-dataset}. On the left plot, we observe that real images (drawn from ImageNet-1k and MS-COCO) have less homogeneous resolutions as they drift away from the diagonal. However, in the right plot, we observe that fake images from AI-generators are practically over the diagonal, with squared resolutions clearly dominating.}
    \label{fig:resolution_confound}
\end{figure*}

\paragraph{The confound.} Public benchmarks as OSSIA, DMID or SynthBuster pair real photographs with generated images that come from very different acquisition pipelines, and image size is one of the most visible differences between the two. Real images are stored at arbitrary resolutions: the 750 ImageNet test images span 248 distinct sizes, and the 750 MS-COCO images 157, and fewer than 5\% \% of them are square. Generated images are, instead, emitted at the handful of resolutions each model was trained to produce. In our test set, 23 of the 26 generators output only square images, and the remaining three (Flux, Midjourney v6, Firefly) use a few fixed aspect ratios. This means that the $256{\times}256$ resolution alone represents half of all fake images. The image size therefore carries the label. To quantify it, we train an XGBoost classifier that sees nothing but the image dimension metadata (i.e., height, width, shorter side, and aspect ratio). The results show a 99.95\% accuracy and 99.96\% AUC averaged over the 26 generators and reaches at least 99.7\% accuracy on each of them. This classifier sees labels and is not a detector, but it shows that a model with access to the native resolution can solve the benchmark without inspecting a single pixel. Fig.~\ref{fig:resolution_confound} shows this visually, where it can be observed that resolution sizes are more heterogeneous in the real images and homogeneous (squared resolutions) in fake samples.

\paragraph{Effect on the detectors.} A detector does not need to be designed around resolution to exploit it: rescaling, tiling, or padding the input makes the network's input statistics depend on the original size, and methods can take advantage of this. We first noticed it in our own model. At native resolution, \method reaches 98.51\% AUC, compared to 97.21\% in center crops: part of the gain comes from the size of the image rather than its content. The effect is much larger for methods whose preprocessing is explicitly dependent on size. This happens explicitly to FSD: it reaches 98.49\% at native resolution and falls to 52.45\% once the cue is removed.

\paragraph{The center-crop protocol.} To remove the shortcut while keeping the forensic content, every method in the main benchmark receives the $256{\times}256$ center crop of each image. The choice of size follows from the data: $256{\times}256$ is the most frequent fake resolution, so half of the fake images are left untouched, and every real and fake test image is at least 256 pixels on each side, so the crop never resamples and no interpolation artifact is introduced. After cropping, all images share the same size and the dimensions no longer carry any information. Table~\ref{tab:resolution} compares all methods before and after removing the confound.

\begin{table*}[t]
\centering
\begin{tabular}{lcccc|cccc|c}
\toprule
& \multicolumn{4}{c|}{Native resolution} & \multicolumn{4}{c|}{Center crop $256{\times}256$} & $\Delta$ \\
Method & GAN & Dif & Com & Global & GAN & Dif & Com & Global & Global Diff. \\
\midrule
CNNDet  & 93.82 & 65.84 & 68.59 & 74.97 & 91.83 & 68.19 & 62.83 & 74.43 & -0.54 \\
UFD     & 97.73 & 82.69 & 85.03 & 87.77 & 97.73 & 82.69 & 85.03 & 87.77 & 0.00 \\
LGrad   & 78.05 & 63.02 & 53.99 & 65.91 & 91.32 & 77.02 & 67.61 & 79.61 & +13.70 \\
NPR     & 92.42 & 94.26 & 74.61 & 89.91 & 90.55 & 89.45 & 68.81 & 85.82 & -4.09 \\
SPAI    & 95.10 & 94.75 & 94.57 & 94.82 & 93.08 & 92.45 & 91.38 & 92.44 & -2.38 \\
EFFORT  & 99.48 & 98.63 & 98.67 & 98.90 & 98.31 & 97.01 & 96.79 & 97.37 & -1.33 \\
OmniAID & 99.54 & 97.23 & 96.04 & 97.71 & 99.40 & 97.66 & 92.30 & 97.17 & -0.54 \\
ConV    & 83.14 & 74.00 & 72.39 & 76.50 & 73.95 & 69.02 & 69.77 & 70.68 & -5.82 \\
FSD     & 99.89 & 99.46 & 93.76 & 98.49 & 57.69 & 52.53 & 43.84 & 52.45 & -46.04 \\
\midrule
Ours    & 98.75 & 99.07 & 96.64 & 98.51 & 97.81 & 97.33 & 95.92 & 97.21 & -1.30 \\
\bottomrule
\end{tabular}
\caption{AUC (\%) at native resolution (left) and on the $256{\times}256$ center crop used in the main benchmark (right), mean over the 26 test generators. Ours: zero-shot, PCA-whitened diagonal Gaussian ($K{=}1$) fitted on real images processed in the same way as the test images. Global Diff. column reflects the difference of performance between the Native Resolution and center crop protocol.}
\label{tab:resolution}
\end{table*}

\paragraph{Results.} Removing the shortcut changes the ranking. Methods with size-dependent preprocessing lose the most (FSD $-46.0$, ConV $-5.8$, NPR $-4.1$), whereas LGrad improves by 13.7 points, so the protocol is not a blanket penalty on any family of methods: it only removes information that is not related to how the image was generated. The UFD is unchanged because its own preprocessing already takes a center crop. Our zero-shot detector is also affected (98.51\% vs.\ 97.21\%), indicating that the density over the forensic representation is mildly dependent on the image size. 

\section{Practical Scenarios}
\label{app:robustness}

\paragraph{Why it matters.} Images found in the wild are rarely the raw output of a generator: social platforms re-compress them, and capture or transmission adds noise. Both operations act on the high-frequency content of the image, which is exactly where low-level forensic detectors look for generation traces. A detector that is accurate on pristine images but collapses after a re-compression is of limited practical use, so we measure how each family of methods degrades.

\paragraph{Protocol.} Each perturbation is applied identically to real and fake test images: JPEG re-compression at quality factors 95, 80, 65 and 50 (Table~\ref{tab:rob_jpeg}) and additive Gaussian noise with $\sigma \in \{0.01, 0.02, 0.05\}$ on $[0,1]$ intensities (Table~\ref{tab:rob_noise}). We compare two families. \emph{Forensic-trace detectors}, which model low-level pixel statistics: NPR and LGrad, evaluated off the shelf, and ours. \emph{VFM-based detectors}, which adapt a large pretrained vision foundation model with real and fake images: EFFORT and OmniAID. To compare representations rather than classifiers, the VFM-based detectors and ours are read through the same two calibrated heads: a k-NN with 25 SD1.4 images and a linear probe trained on 1{,}000 SD1.4 images. Both heads are fitted on native and JPEG-compressed calibration images, and their hyper-parameter ($k$ or $C$) is selected on validation. Two consequences follow: JPEG is a condition the heads have seen during calibration, whereas noise is not, so Table~\ref{tab:rob_noise} is the cleaner test of how robust each representation is by itself; and EFFORT was trained on SD1.4, the calibration generator.

\begin{table}[t]
\centering
\resizebox{\linewidth}{!}{%
\begin{tabular}{llccccc}
\toprule
Method & Head & Native & QF95 & QF80 & QF65 & QF50 \\
\midrule
\multicolumn{7}{l}{\textit{Forensic-trace detectors}} \\
NPR     & off-the-shelf & 85.82 & 30.02 & 34.09 & 34.50 & 35.24 \\
LGrad   & off-the-shelf & 79.61 & 51.50 & 60.99 & 58.18 & 60.20 \\
Ours    & zero-shot     & 97.21 & 63.45 & 62.00 & 58.36 & 50.76 \\
Ours    & k-NN          & 95.40 & 78.65 & 79.12 & 74.83 & 72.63 \\
Ours    & probe         & 95.10 & 83.80 & 84.91 & 81.85 & 76.78 \\
\midrule
\multicolumn{7}{l}{\textit{VFM-based detectors (trained on fakes)}} \\
EFFORT  & k-NN          & 96.13 & 89.77 & 85.82 & 78.58 & 77.61 \\
OmniAID & k-NN          & 98.07 & 94.26 & 90.16 & 83.52 & 84.63 \\
EFFORT  & probe         & 98.74 & 97.03 & 95.29 & 92.11 & 91.34 \\
OmniAID & probe         & 98.69 & 96.73 & 95.56 & 92.71 & 91.91 \\
\bottomrule
\end{tabular}}
\caption{AUC (\%) under JPEG re-compression at quality factor QF, mean over the 26 test generators. Calibrated heads use 25 (k-NN) or 1{,}000 (probe) SD1.4 images, fitted on native and JPEG-compressed calibration images.}
\label{tab:rob_jpeg}
\end{table}

\begin{table}[t]
\centering
\resizebox{0.95\linewidth}{!}{%
\begin{tabular}{llcccc}
\toprule
Method & Head & Native & $\sigma{=}0.01$ & $\sigma{=}0.02$ & $\sigma{=}0.05$ \\
\midrule
\multicolumn{6}{l}{\textit{Forensic-trace detectors}} \\
NPR     & off-the-shelf & 85.82 & 44.59 & 36.26 & 34.45 \\
LGrad   & off-the-shelf & 79.61 & 56.30 & 48.59 & 48.16 \\
Ours    & zero-shot     & 97.21 & 83.56 & 82.08 & 78.72 \\
Ours    & k-NN          & 95.40 & 80.58 & 78.67 & 62.53 \\
Ours    & probe         & 95.10 & 87.97 & 84.61 & 79.54 \\
\midrule
\multicolumn{6}{l}{\textit{VFM-based detectors (trained on fakes)}} \\
EFFORT  & k-NN          & 96.13 & 86.56 & 83.79 & 80.75 \\
OmniAID & k-NN          & 98.07 & 94.53 & 91.35 & 83.28 \\
EFFORT  & probe         & 98.74 & 94.46 & 92.19 & 88.08 \\
OmniAID & probe         & 98.69 & 95.68 & 93.67 & 87.30 \\
\bottomrule
\end{tabular}
}%
\caption{AUC (\%) under additive Gaussian noise of standard deviation $\sigma$ (intensities in $[0,1]$), mean over the 26 test generators. No noise is seen when fitting any of the heads.}
\label{tab:rob_noise}
\end{table}

\paragraph{Forensic-trace detectors break under post-processing.} Detectors that rely on low-level generation traces degrade sharply once the image is re-encoded. QF95 alone brings NPR from 85.82\% to 30.02\%, below chance at every quality factor and noise level, and LGrad loses about 20 points under JPEG and falls to chance under noise. Our zero-shot detector follows the same trend under compression (63.45\% at QF95, 50.76\% at QF50), since JPEG removes much of the high-frequency residual on which the density is fitted. Noise is the exception: it perturbs the residual without erasing it, and the zero-shot detector keeps 78.72\% at $\sigma{=}0.05$, 30 to 45 points above NPR and LGrad. Calibrating with JPEG-compressed images raises the mean JPEG AUC to 81.83\% with the linear probe.

\paragraph{Why VFM-based detectors are more robust.} EFFORT and OmniAID degrade far more gracefully: with a linear probe, both stay above 91\% at QF50 and 87\% at $\sigma{=}0.05$. We see two non-exclusive reasons. \emph{First, they are trained on fake images}, with augmentation in the case of EFFORT, so their representation learns to keep the real/fake distinction while discarding these perturbations; ours never sees a fake and preserves precisely the residual statistics these operations destroy. \emph{Second, their representation is largely semantic} (Sec.~\ref{sec:forensic_representation_analysis}): a 1-NN on seven ImageNet super-classes reaches 99.3\% with OmniAID features against 39.0\% with ours, and content is almost invariant to re-compression and mild noise. Since no VFM-based detector is trained without fakes, we cannot separate the two factors. Robustness to post-processing thus remains the main limitation of a representation learned from real images alone.

\section{Computational Cost}
\label{app:complexity}

\begin{table*}[t]
\centering
\begin{tabular}{@{}llrrrr@{}}
\toprule
\textbf{Method} & \textbf{Inference Type} & \textbf{\#Params} & \textbf{GFLOPs}$\downarrow$ & \textbf{Lat. (ms)}$\downarrow$ & \textbf{GPU (MB)}$\downarrow$ \\
\midrule
CNNDet~\cite{wang2020cnn} & Forward & 23.5M & 10.7 & 2.94 & 608 \\
UFD & Forward & 304.0M & 162.0 & 8.14 & 2176 \\
LGrad & Forward + backward & 46.6M & 102.0 & 12.50 & 1006 \\
SPAI & Forward & 139.9M & 656.8 & 26.42 & 1548 \\
NPR & Forward & 1.4M & 4.5 & 1.61 & 518 \\
FSD & Per-image solve & 6.2M & 2537.0 & 2413.66 & 11342 \\
ConV & Forward (10 crops) & 304.4M & 1620.2 & 16.82 & 1176 \\
EFFORT & Forward & 504.6M & 162.2 & 10.21 & 2448 \\
OmniAID & Forward & 508.8M & 765.7 & 40.64 & 3672 \\
\midrule
Ours (zero-shot, GMM) & Forward & 3.0M & 10.7 & 5.73 & 516 \\
Ours (few-shot, $k$-NN) & Forward & 3.0M & 10.7 & 5.37 & 516 \\
Ours (linear probe) & Forward & 3.0M & 10.7 & 5.60 & 516 \\
\bottomrule
\end{tabular}
\caption{Computational cost of inference on a fixed sample of 500 ImageNet-1k images, center-cropped to $256\times256$ as in the detection benchmark (same images for every row). GFLOPs and latency start from an already decoded and preprocessed input; latency is the per-image median on one RTX 4090; memory is the peak GPU footprint over the model's lifetime.}
\label{tab:computational_efficiency}
\end{table*}

\textbf{Computational Cost Assessment.} Table~\ref{tab:computational_efficiency} reports the inference cost of every detector. In terms of parameters, \method (3.0M) is the second smallest model, behind only NPR (1.4M) and ahead of FSD (6.2M). It is at least $100\times$ smaller than the detectors built on foundation backbones (UFD, ConV, EFFORT and OmniAID, 304--509M parameters), and it is trained on fewer than 0.5M real images, yet it outperforms UFD and performs on par with OmniAID and EFFORT (Table~\ref{tab:det_results_full}). In terms of latency, NPR (1.61\,ms) and CNNDet (2.94\,ms) are the fastest detectors, followed by \method (5.73\,ms in zero-shot, 5.37\,ms in few-shot and 5.60\,ms in linear probe), which is $~1.4\times$ faster than UFD, $1.8\times$ faster than EFFORT, $2.9\times$ faster than ConV, and $7.1\times$ faster than OmniAID. \method has the second lowest cost of all detectors in GFLOPs (10.7, the same as CNNDet and behind 4.5 for NPR) and in GPU memory (516\,MB, similar to NPR and CNNDet). FSD is the clear outlier: its per-image solve makes it the most expensive detector in GFLOPs (2537.0), latency (2,413.66\,ms, $421\times$ slower than \method) and memory (11,342\,MB), which severely limits its use in production settings. Overall, \method and NPR are the only detectors that rank among the three cheapest on all four cost metrics.

\section{Robustness Against Unseen Real Images}
\label{app:unseen_real}
Since the \ac{gmm} is fitted to real image embeddings alone, its coverage of the real domain in test-time matters. We retrain the encoder on a single real source and fit the \ac{gmm} on that same source alone, so the other source is never observed at any stage, and then evaluate both real axes. Against the main model ---95.98\% AUC on ImageNet and 98.44\% on MS-COCO--- the ImageNet-only scenario (500k images) drops the in-distribution ImageNet axis to 94.82\% and the now-unseen MS-COCO axis to 91.82\%, for a global mean of 93.32\%. The MS-COCO-only scenario behaves differently despite training on 242k images, less than half the corpus: the unseen ImageNet axis gives up 4.41\% at 91.57\%, while its own axis rises to 98.78\%, for a global mean of 95.18\%. Detection therefore survives a wholly unseen real domain in both directions, but the cost is asymmetric: removing MS-COCO costs 3.89\% globally, whereas removing ImageNet costs 2.03\% with less than half the training images. The unseen axis is more sensitive to the density model: with a full coefficient \ac{gmm}, which does not whiten the embedding, it rises to 93.70\% (MS-COCO) and 95.22\% (ImageNet), suggesting that whitening amplifies low-variance directions along which real domains differ.

\section{Detailed Breakdown on Generators}
\label{app:detailed_generators}

\paragraph{Overview.}
The detection benchmark comprises 26 generator families and two real-image distributions. Table~\ref{tab:generators_breakdown} describes every family as it enters the evaluation: the architecture and conditioning of the generator, the domain of content it was trained on or prompted with, the resolution and file format of the evaluated images and the number of images used. Images are drawn from three public collections of synthetic images, DMID~\citep{dmid-dataset}, SynthBuster~\citep{synthbuster-dataset} and OSSIA~\citep{ossia-dataset}; the sub-sets merged into each family are listed in the last column.

\begin{table*}
\centering
\scriptsize
\setlength{\tabcolsep}{3pt}
\resizebox{\textwidth}{!}{%
\begin{tabular}{@{}l r r l@{}}
\toprule
\textbf{Family} & \textbf{Resolution} & \textbf{$N$} & \textbf{Sources} \\
\midrule
\multicolumn{4}{@{}l}{\textit{Generative Adversarial Networks}} \\
ProGAN  & $256^2$ (67\%), $1024^2$ (33\%) & 750 & DMID, SynthBuster, OSSIA \\
StyleGAN2 & $1024^2$ (45\%), $256^2$ (42\%), $512^2$ (13\%) & 750 & DMID, SynthBuster, OSSIA \\
StyleGAN3     & $1024^2$ (56\%), $256^2$ (28\%), $512^2$ (15\%)    & 750 & DMID, SynthBuster, OSSIA \\
BigGAN        & $256^2$ (60\%), $512^2$ (40\%)                     & 750 & DMID, SynthBuster, OSSIA \\
StarGAN       & $256^2$                              & 750 & SynthBuster, OSSIA \\
EG3D          & $512^2$                                            & 750 & DMID \\
GauGAN        & $256^2$                                            & 750 & SynthBuster \\
GigaGAN       & $256^2$ (50\%), $512^2$ (50\%)                     & 750 & SynthBuster \\
\midrule
\multicolumn{4}{@{}l}{\textit{Diffusion and autoregressive models}} \\
DDPM          & $256^2$                                            & 750 & SynthBuster, OSSIA \\
Guided Diff.\ (ADM) & $256^2$ & 750 & DMID, SynthBuster \\
LDM           & $256^2$ & 750 & DMID, SynthBuster, OSSIA \\                 
Stable Diffusion 2.1 & $512^2$ (40\%), $256^2$ (30\%), $768^2$ (30\%)     & 750 & SynthBuster \\
Stable Diffusion 3 & $1024^2$                                           & 750 & OSSIA \\
SDXL          & $1024^2$                                           & 750 & SynthBuster \\
Flux          & $1024\times768$, $768\times1024$, $1024^2$         & 750 & Online \\
GLIDE         & $256^2$                                            & 750 & DMID, SynthBuster \\
DeepFloyd IF  & $1024^2$                                           & 750 & SynthBuster \\
DiT           & $256^2$ (50\%), $512^2$ (50\%)                     & 750 & SynthBuster \\
Diffusion-GAN & $256^2$                                            & 750 & SynthBuster \\
LSGM          & $256^2$                                            & 750 & OSSIA \\
Taming Transformers & $256^2$                                      & 750 & DMID \\
\midrule
\multicolumn{4}{@{}l}{\textit{Commercial / black-box APIs}} \\
Midjourney v6 & Variable aspect ratio, shorter side 260--2048      & 750 & Online \\
Adobe Firefly & $512\times427$ (80\%), $512^2$ (15\%), $672\times299$ (5\%) & 750 & Online \\
DALL$\cdot$E~2 & $1024^2$                                           & 750 & DMID, SynthBuster \\
DALL$\cdot$E~3 & $1024^2$                                           & 750 & SynthBuster \\
DALL$\cdot$E Mini & $256^2$                                         & 750 & DMID \\
\midrule
\multicolumn{4}{@{}l}{\textit{Real images}} \\
ImageNet-1k   & Variable, median shorter side 375                  & 750 & ILSVRC2012 \\
MS-COCO       & Variable, median shorter side 428                  & 750 & MS-COCO 2017 \\
\bottomrule
\end{tabular}%
}
\caption{Generator families of the detection benchmark. Resolution is width$\times$height of the evaluated images; the proportion of each resolution or content domain within the family is given in parentheses. $N$ is the number of evaluated images. }
\label{tab:generators_breakdown}
\end{table*}

\paragraph{Diversity of the benchmark.}
The families cover the three main generation paradigms (adversarial, diffusion, and autoregressive), and resolutions range from $256\times256$ to $2048$ pixels on the shorter side, and several families mix resolutions within themselves (ProGAN, StyleGAN2, StyleGAN3, BigGAN, GigaGAN, Stable Diffusion 2.1, DiT, Flux, Midjourney), so a detector cannot rely on a single fixed image size to recognize a generator. For these families, the generated images depict the same kind of scenes as the MS-COCO real images, so the MS-COCO comparison cannot be solved on content alone, whereas the ImageNet-1k comparison additionally differs in content; this is one more reason to report both real distributions.
\end{document}